\documentclass[conference]{IEEEtran}

\IEEEoverridecommandlockouts

\usepackage{amsmath}
\usepackage{graphicx}
\usepackage{tikz}
\usetikzlibrary{arrows.meta,calc}

\definecolor{Vnavy}{HTML}{081A58}
\definecolor{Vline}{HTML}{476782}
\definecolor{Vblue}{HTML}{159CFF}
\definecolor{Vteal}{HTML}{00B5B0}
\definecolor{Vdarkteal}{HTML}{006B79}
\definecolor{Vaqua}{HTML}{77D7DC}
\definecolor{Vpurple}{HTML}{9475ED}
\definecolor{Vorange}{HTML}{FF821E}
\definecolor{Vgold}{HTML}{FFC454}
\definecolor{Vpink}{HTML}{ED72C2}
\definecolor{Vred}{HTML}{F52232}

\newcommand{\VHeaderBox}[7]{%
  \begin{scope}
    \clip[rounded corners=6pt]
      (#1,#2) rectangle (#3,#4);
    \fill[white] (#1,#2) rectangle (#3,#4);
    \fill[#5!13]
      (#1,{#4-#6}) rectangle (#3,#4);
  \end{scope}
  \draw[draw=#5,line width=#7,rounded corners=6pt]
    (#1,#2) rectangle (#3,#4);
}

\newcommand{\VToken}[5]{%
  \shade[
    draw=#5,line width=.45pt,
    top color=#5!28,bottom color=#5!65
  ] ({#1},{#2}) rectangle ++({#3},{#4});
}

\usepackage{cite}
\usepackage{amsmath,amssymb,amsfonts}
\usepackage{algorithmic}
\usepackage{graphicx}
\usepackage{textcomp}
\usepackage{xcolor}
\usepackage{url}
\usepackage{placeins}

\usepackage{wrapfig}   
\usepackage{colortbl}  

\newcommand{\name}{{\tt VLA-Scope}}

\definecolor{mygray}{gray}{0.92}
\providecommand{\thickhline}{%
    \noalign{\hrule height 0.8pt}%
}
\newif\ifshowreview
\showreviewtrue
\newif\ifshowrevisions
\showrevisionstrue

\def\BibTeX{{\rm B\kern-.05em{\sc i\kern-.025em b}\kern-.08em
    T\kern-.1667em\lower.7ex\hbox{E}\kern-.125emX}}

\begin{document}
\bstctlcite{BSTcontrol}

\title{\name{}: Shift-Aware Failure Prediction for Vision-Language-Action Models}

\author{
\IEEEauthorblockN{
Kaiwen Zhu\IEEEauthorrefmark{1},
Dongfang Liu\IEEEauthorrefmark{2},
and Liangkai Liu\IEEEauthorrefmark{1}
}
\IEEEauthorblockA{\IEEEauthorrefmark{1}
Department of Computer Science, Texas Tech University, Lubbock, TX, USA}
\IEEEauthorblockA{\IEEEauthorrefmark{2}
Purdue University, West Lafayette, IN, USA}
}

\maketitle

\begin{abstract}
Vision-language-action (VLA) models map visual observations and natural-language instructions to robotic actions, but distribution shifts can compromise their reliability. Because these models may still succeed under out-of-distribution (OOD) conditions, detecting OOD inputs alone is insufficient to predict execution failure. In this paper, we introduce \name{}, a two-stage framework that combines input-shift characterization with execution history to predict failure during OOD rollouts. The first stage uses pooled image and language representations to detect OOD inputs and classify their shift categories. For inputs flagged as OOD, the second stage combines the predicted category, action-prefix features, and execution progress features. A logistic regression model shared across shift categories updates failure risk as execution proceeds. We evaluate the framework with OpenVLA on ten LIBERO-Spatial tasks using leave-one-group-out cross-validation. OOD detection achieves a ROC-AUC of 0.9454, and shift classification achieves 91\% accuracy. Evaluated independently of the OOD gate on all 1,400 OOD rollouts, the failure predictor achieves a ROC-AUC of 0.8497 after 60 executed actions, compared with 0.7906 without execution progress features. It also achieves higher ROC-AUC than the evaluated ActProbe and SAFE-MLP baselines. These results suggest that combining action features with temporally aggregated execution-step representations improves failure prediction under input shifts.
\end{abstract}



\section{Introduction}
Vision-language-action (VLA) models adapt pretrained vision-language representations to map visual observations and natural-language instructions to robotic actions~\cite{rt2,openvla,pi0}. Large-scale demonstration datasets support learning across robots and tasks~\cite{openxembodiment}. During deployment, however, changes in camera viewpoint, visual appearance, instructions, object layout, or robot initialization can shift inputs away from reference conditions~\cite{colosseum,liberoplus}.

\begin{figure}[!t]
    \centering
    \includegraphics[width=\columnwidth]{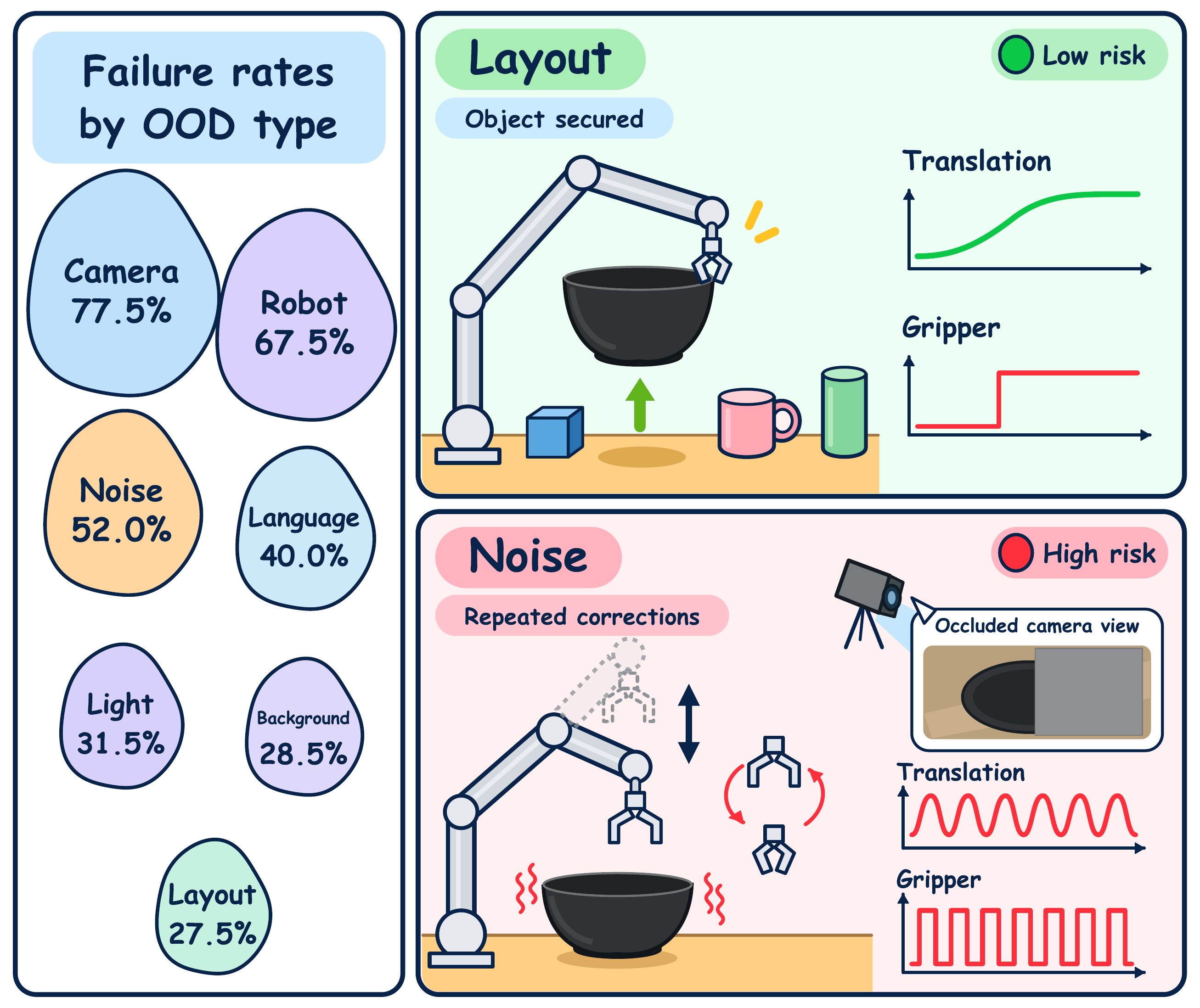}
    \caption{OOD inputs do not always lead to failure. Left: observed failure rates across seven OOD categories. Right: illustrative examples of a secure grasp after a layout change and repeated corrections under visual noise. Robot motion and gripper behavior provide additional clues for predicting failure.}
    \label{fig:intro_motivation}
\end{figure}

Out-of-distribution (OOD) detection distinguishes unfamiliar inputs from an in-distribution (ID) reference using, for example, classifier scores or feature-space distances~\cite{oodsurvey,hendrycks2017baseline,energyood,mahalanobis}. However, a binary OOD label neither identifies the shift type nor establishes whether execution will fail. A VLA policy may succeed under a shift, and rollouts within the same OOD category may have different outcomes. Task-driven OOD detection explicitly considers violations of policy performance bounds~\cite{taskdrivenood}. Our setting separately characterizes the initial input shift and predicts the eventual rollout outcome. Figure~\ref{fig:intro_motivation} summarizes observed failure rates across OOD categories and illustrates execution-time action patterns, motivating the use of both shift context and execution evidence for failure prediction.

Runtime monitors predict failure risk from observations and action behavior~\cite{sentinel,fiper,faildetect} or internal policy representations~\cite{safe,safecast,earlywarning}. ActProbe, in particular, uses action magnitude and temporal consistency in a recurrent predictor conditioned on an instruction embedding~\cite{actprobe2026}. These approaches motivate examining how action behavior and internal representations complement each other under input shifts. Action features summarize the commands issued so far, whereas execution-step representations encode the visual and language context used to generate actions. We investigate their combination with a shift category inferred from the initial input to predict failure from partial executions.

We introduce \name{}, a two-stage diagnostic framework for VLA models. The first stage, the \emph{OOD Characterizer}, uses pooled image and language representations from the initial input to distinguish ID from OOD inputs. A separate classifier assigns inputs identified as OOD to predefined shift categories. The second stage, the \emph{Failure Risk Predictor}, combines the predicted category with 14 action-prefix features and a compact, temporally aggregated representation. The features cover per-axis translation and rotation magnitudes, adjacent-action changes, and gripper commands. At each execution step, we extract the final-layer hidden vector at the last prompt token position during action generation. We cumulatively average these vectors from step 10 to the current step and compress the average to 32 dimensions using training-fitted standardization and principal component analysis. The predicted category remains fixed within a rollout, while the action features and cumulative representation are updated during execution.

A shared logistic regression model learns from these prefix-level inputs using final rollout outcomes as supervision. The same model serves all OOD categories without modifying the underlying VLA policy or requiring an ID action reference. Its linear score separates the contributions of action features and the compressed cumulative representation.

We instantiate the framework with OpenVLA~\cite{openvla} as a representative VLA model. We evaluate the diagnostic models offline using leave-one-group-out cross-validation on ten LIBERO-Spatial tasks~\cite{libero}, holding out an entire task in each fold. Our evaluation includes 500 ID inputs and 1,400 OOD rollouts across seven shift categories, generated using LIBERO-Plus perturbation categories and assets~\cite{liberoplus} under our own sampling and configuration protocol. OOD detection achieves a ROC-AUC of 0.9454, and shift classification achieves 91\% accuracy. Evaluated independently of the binary OOD gate on all 1,400 OOD rollouts, failure prediction achieves a ROC-AUC of 0.8497 after 60 executed actions, compared with 0.7906 without the representation input. At this checkpoint, the predictor also achieves higher ROC-AUC than the evaluated ActProbe and SAFE-MLP baselines.

The main technical contributions are:
\begin{itemize}
\item \textbf{Two-stage diagnosis from input shift to task outcome.}
We connect initial ID/OOD detection and shift-type classification with sequential failure-risk prediction, separating input characterization from the likelihood of task failure.

\item \textbf{Failure prediction from actions and accumulated representations.}
We develop a shared logistic risk predictor that combines inferred shift context, explicit action-prefix features, and cumulatively averaged execution-step representations to assess partial executions.

\item \textbf{Evaluation of prediction accuracy and temporal stability.}
We compare representation aggregation strategies on held-out tasks, examine risk fluctuations and alarm behavior, and compare ranking performance against existing failure predictors.
\end{itemize}

\section{Related Work}
\subsection{Vision-Language-Action Models}

Vision-language-action models map visual observations and language instructions to robot actions, using discretized commands~\cite{rt2,openvla}, flow matching~\cite{pi0}, or frequency-domain action tokenization~\cite{fast}. Open X-Embodiment provides demonstrations across robots and tasks~\cite{openxembodiment}. LIBERO evaluates task transfer~\cite{libero}, while Colosseum and LIBERO-Plus assess robustness to perturbations~\cite{colosseum,liberoplus}. Li et al.~\cite{vlageneralizable} adapt visual representations to improve viewpoint robustness. We instead study initial input shifts and execution-time failure risk around a frozen OpenVLA policy.

\subsection{Out-of-Distribution Detection}

OOD detection distinguishes inputs departing from a reference distribution~\cite{oodsurvey}. Common scores use maximum softmax probabilities~\cite{hendrycks2017baseline}, energy~\cite{energyood}, or distances to class-conditional feature distributions~\cite{mahalanobis}. In robotics, Farid et al.~\cite{taskdrivenood} connect distribution shift to violations of policy performance bounds. Our setting separates initial input-shift characterization from execution-outcome prediction: supervised classifiers detect OOD inputs and identify seven predefined perturbation types, providing context for a separate failure-risk predictor.

\subsection{Failure Detection and Prediction}

Sentinel~\cite{sentinel} combines action-consistency monitoring with vision-language assessment of task progress. FIPER~\cite{fiper} combines observation novelty and action-chunk entropy, calibrating scores on successful rollouts. FAIL-Detect~\cite{faildetect} treats failure detection as sequential OOD detection, using successful rollouts to calibrate time-varying thresholds through conformal prediction.

Internal-feature methods include SAFE~\cite{safe}, which trains MLP and LSTM detectors using rollout outcomes, and SAFECAST~\cite{safecast}, which adds visual and language contrast sets to probe training and calibration for deployment shifts. Mahato and Ren~\cite{earlywarning} train linear monitors on frozen OpenVLA activations to identify steps preceding logged failure events. Our target is instead the eventual rollout outcome.

ActProbe~\cite{actprobe2026} predicts outcomes with a task-instruction-conditioned recurrent model using action magnitude and temporal consistency. For OpenVLA, consistency uses squared differences between consecutive actions. We combine explicit action-prefix features with a compact representation obtained by cumulatively averaging execution-step representations and an OOD category predicted from the initial observation and instruction. A shared logistic regression model uses these inputs to predict eventual failure. Its linear form separates contributions from action features and the compressed cumulative representation.

\section{\name{} Design}
\subsection{System Overview}

\begin{figure*}[t]
    \centering
    \resizebox{\textwidth}{!}{
    \input{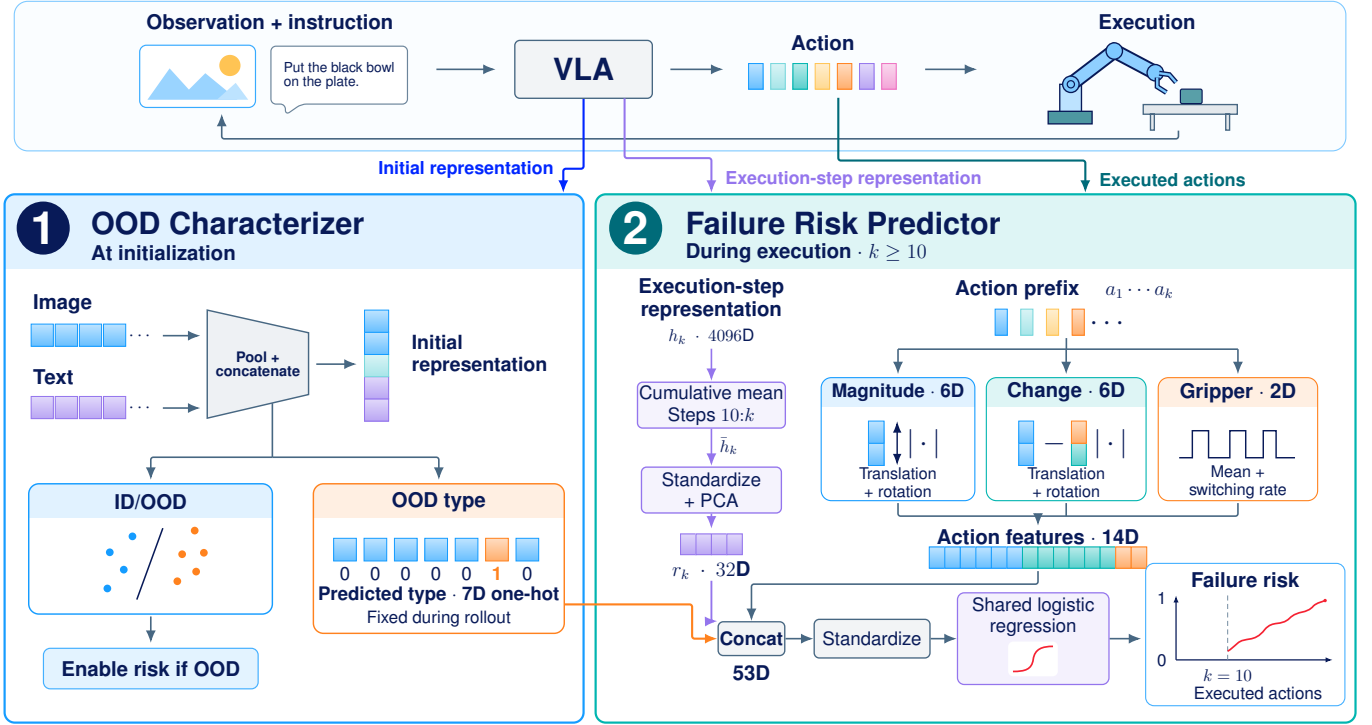}
    }
    \caption{Overview of \name{}.
Module 1 detects initial input shifts and predicts their types.
For OOD inputs, Module 2 combines 14 action-prefix features,
a seven-dimensional predicted-type encoding, and a
32-dimensional representation obtained by cumulatively averaging
execution-step representations from step 10 and applying PCA.
The resulting 53-dimensional vector is standardized and
passed to a shared logistic regression model to update
failure risk. The risk curve is schematic.}
    \label{fig:vla_ood_system}
\end{figure*}

\name{} is a general diagnostic framework for VLA models, as shown in Fig.~\ref{fig:vla_ood_system}. The VLA model maps a camera image with a language instruction to a decoded seven-dimensional action command, comprising translation, rotation, and gripper components. The robot executes the action and returns the next observation. \name{} reads signals from this process without changing the model weights or its actions.

The framework is designed with two stages: an initial input OOD characterizer and an execution-time risk predictor. The OOD Characterizer uses the initial observation and instruction to make an ID/OOD decision and, for inputs classified as OOD, predict an OOD type. The Failure Risk Predictor combines this predicted category with features of the actions executed so far and a cumulative average of execution-step representations obtained during action generation. The category remains fixed throughout the rollout, whereas the action features, cumulative representation, and risk score are updated during execution.

Let $\boldsymbol{a}_{1:k}$ denote the decoded commands corresponding to the first $k$ executed policy actions. The prediction target is the eventual rollout outcome: $y=0$ if the environment reports task success within the action budget, and $y=1$ otherwise. The predictor outputs a score in $[0,1]$ for this final failure outcome, using the action-prefix features, predicted OOD type, and representations accumulated from step 10 through step $k$.

\subsection{OOD Characterizer}

We propose to leverage the multimodal representation of the initial observation ($\boldsymbol{z}_0\in\mathbb{R}^{8192}$) to predict potential ID/OOD types. This can be obtained from token embeddings immediately before any policy action is executed. Image tokens and valid text tokens are mean-pooled separately, excluding the beginning-of-sequence token. The resulting two 4096-dimensional vectors are concatenated to form $\boldsymbol{z}_0$. The subscript denotes the initial time. 

Two separately trained logistic regression classifiers use $\boldsymbol{z}_0$. Their prediction functions include their respective preprocessing, fitted only on the corresponding training data. Both classifiers use class-balanced training weights. The binary classifier outputs the predicted probability of the OOD class, which we use as the OOD score $d(\boldsymbol{z}_0)\in[0,1]$.

Let $\mathcal{C}$ contain the seven OOD types: \texttt{Background}, \texttt{Camera}, \texttt{Language}, \texttt{Light}, \texttt{Noise}, \texttt{Layout}, and \texttt{Robot}. The type classifier outputs class probabilities $q_c(\boldsymbol{z}_0)$ for $c\in\mathcal{C}$ and selects $\hat c=\arg\max_{c\in\mathcal{C}}q_c(\boldsymbol{z}_0)$. The risk module encodes this predicted category as $\boldsymbol{e}(\hat c)\in\mathbb{R}^{7}$. The type classifier takes the initial representation as input, rather than the binary classifier's score.

We use the binary ID/OOD decision as a gate for failure-risk prediction. The threshold is shared across task groups, and the risk predictor is enabled only when
\begin{equation}
    d(\boldsymbol{z}_0)>\tau_{\mathrm{OOD}},
    \qquad \tau_{\mathrm{OOD}}=0.5.
    \label{eq:ood_gate}
\end{equation}
When enabled, the gate passes $\hat c$ to the Failure Risk Predictor. Otherwise, the system returns an ID decision without a failure-risk score; this is not a prediction of task success. The OOD Characterizer operates only at initialization, so the predicted category is not updated for shifts arising later.

\subsection{Failure Risk Predictor}

\vspace{1mm}
\noindent\textbf{Action commands.}
We use the decoded command associated with each executed policy action, recorded before simulator-specific transformations. Its components are ordered as
\[
    \boldsymbol{a}_i=
    [t_{i,x},t_{i,y},t_{i,z},r_{i,x},r_{i,y},r_{i,z},g_i]^\top,
\]
where the first three components control translation, the next three control rotation, and $g_i$ controls the gripper. The policy maps each action token to a discrete bin center $u_{i,j}$, clipping the bin index to its valid range. For the first six components, it restores the command scale using the action features stored in the checkpoint:
\begin{equation}
    a_{i,j}=q_{01,j}
    +\frac{u_{i,j}+1}{2}(q_{99,j}-q_{01,j}),
    \quad j=1,\ldots,6,
    \label{eq:action_unnormalization}
\end{equation}
where $q_{01,j}$ and $q_{99,j}$ are the stored first and ninety-ninth percentiles. The checkpoint excludes the gripper component from this transformation, so $g_i=u_{i,7}$.

In our design, the operational-space pose controller clips the first six components to $[-1,1]$, then scales translation by $0.05$ and rotation by $0.5$. The resulting translation is a world-frame position increment in meters. The rotation is an axis-angle increment in radians, whose rotation matrix left-multiplies the current world-frame end-effector orientation. Our features use the decoded values before this clipping and scaling; the feature extractor applies no additional clipping.

The runtime converts the gripper command to $g_i^{\mathrm{sim}}=-\operatorname{sign}(2g_i-1)$ before execution. Negative simulator commands open the gripper and positive commands close it. The risk model retains the decoded value $g_i$ and measures switching between positive and nonpositive decoded commands.

\vspace{1mm}
\noindent\textbf{Action features.}
For each translation or rotation component $j=1,\ldots,6$, we compute its mean absolute magnitude $m_{k,j}$ and mean adjacent-command change $v_{k,j}$ over the available prefix. We also compute the mean signed gripper command $\bar g_k$ and its switching rate $\rho_k$:
\begin{equation}
\begin{aligned}
    m_{k,j} &= \frac{1}{k}\sum_{i=1}^{k}|a_{i,j}|,\\
    v_{k,j} &= \frac{1}{k-1}\sum_{i=2}^{k}|a_{i,j}-a_{i-1,j}|,\\
    \bar g_k &= \frac{1}{k}\sum_{i=1}^{k}g_i,\\
    \rho_k &= \frac{1}{k-1}\sum_{i=2}^{k}
    \mathbf{1}\!\left[(g_i>0)\ne(g_{i-1}>0)\right].
\end{aligned}
    \label{eq:action_statistics}
\end{equation}
Here, $\mathbf{1}[\cdot]$ is the indicator function. There are $k$ commands and $k-1$ adjacent pairs. The 14-dimensional feature vector follows the implementation order:
\begin{equation}
    \boldsymbol{s}_k=
    [\bar g_k,m_{k,1},\ldots,m_{k,6},
    v_{k,1},\ldots,v_{k,6},\rho_k]^\top.
    \label{eq:action_feature_vector}
\end{equation}
These features summarize decoded commands and retain separate values for each translation and rotation axis.

\vspace{1mm}
\noindent\textbf{Cumulative execution-step representation.}
For each executed action $i$, we retain
$\boldsymbol{h}_i\in\mathbb{R}^{4096}$, the final-layer hidden vector at the last prompt token position from the forward pass that generated that action. This vector is computed from the observation preceding the action and the language instruction.

At scoring step $k$, we average the representations from step 10 through the current step:
\begin{equation}
    \bar{\boldsymbol{h}}_k
    =\frac{1}{k-9}\sum_{i=10}^{k}\boldsymbol{h}_i,
    \qquad k\geq10.
    \label{eq:cumulative_representation}
\end{equation}
Thus, $\bar{\boldsymbol{h}}_{10}=\boldsymbol{h}_{10}$, and subsequent updates incorporate only representations already available during execution. The cumulative average aggregates visual and language context across the observed execution history.

To reduce the dimensionality gap between the 4096-dimensional representation and the 14 action features, we standardize the cumulative average and apply principal component analysis (PCA), retaining 32 components:
\begin{equation}
    \boldsymbol{r}_k=
    P_{32}\!\left(S_h(\bar{\boldsymbol{h}}_k)\right)
    \in\mathbb{R}^{32},
    \label{eq:execution_representation}
\end{equation}
where $S_h$ is a feature-wise standardizer and $P_{32}$ is a PCA transform, including centering. Both are fitted on cumulative representations from the training groups of each split. To limit preprocessing cost, we use up to eight evenly spaced eligible prefixes per training rollout to fit these transforms. The risk classifier uses all eligible prefixes.

\vspace{1mm}
\noindent\textbf{Risk prediction.}
We concatenate the predicted OOD type, action features, and compressed cumulative representation:
\begin{equation}
    \boldsymbol{x}_k=
    [\boldsymbol{e}(\hat c)^\top,
     \boldsymbol{s}_k^\top,
     \boldsymbol{r}_k^\top]^\top
    \in\mathbb{R}^{53}.
    \label{eq:risk_input}
\end{equation}
A standardization transform $S_x$, fitted on the corresponding training inputs, is applied to this vector. A shared logistic regression model computes the logit $\ell_k$ and failure-risk score $\hat p_k$:
\begin{equation}
    \ell_k=b+\boldsymbol{w}^\top S_x(\boldsymbol{x}_k),
    \qquad
    \hat p_k=\frac{1}{1+\exp(-\ell_k)},
    \label{eq:risk_score}
\end{equation}
where $b$ is the learned intercept and
$\boldsymbol{w}\in\mathbb{R}^{53}$ is the coefficient vector.
We retain all seven type indicators and apply an $L_2$ penalty to $\boldsymbol{w}$. The same parameters are used across scoring steps and OOD types. The predicted type contributes an additive offset, while the action and representation coefficients are shared across types.

\noindent\textbf{Sequential training and scoring.}
The action features are defined for $k\geq2$; our main configuration starts scoring at $k_{\min}=10$. For a rollout terminating after $T$ policy actions, training uses every prefix $k=10,\ldots,T-1$, supervised by the final outcome $y$. Each prefix receives weight $1/(T-10)$, giving each rollout equal total training weight. These weights are used when fitting $S_x$ and the logistic regression model. The counts $k$ and $T$ exclude the initial settling actions. 

Training and scoring use the same cumulative representation construction. During execution, each scoring call computes features over actions $1$ through $k$, averages the retained representations from steps $10$ through $k$, and applies the fitted transforms and shared classifier. The predicted OOD type and preprocessing parameters remain fixed. The classifier output is used directly without additional temporal smoothing; the final scored prefix is $T-1$.

\noindent\textbf{Feature contributions.}
Within a rollout, the predicted type, intercept, and preprocessing parameters remain fixed. For scoring steps $u$ and $v$, the logit difference separates into action-statistic and cumulative-representation contributions:
\begin{equation}
\begin{aligned}
    \ell_v-\ell_u
    ={}&\sum_{j=1}^{14}
    \frac{w_{7+j}}{\sigma_{7+j}}
    (s_{v,j}-s_{u,j})\\
    &+\sum_{j=1}^{32}
    \frac{w_{21+j}}{\sigma_{21+j}}
    (r_{v,j}-r_{u,j}),
\end{aligned}
    \label{eq:logit_difference}
\end{equation}
where $\sigma_j$ is the training-fitted scale of input dimension $j$ in $S_x$. The first sum gives contributions from individual action features; the second gives contributions from the compressed cumulative representation.

\section{Evaluation}
\subsection{Experimental Setup}

\setcounter{dbltopnumber}{2}
\noindent\textbf{Environment and data.}
We evaluate a frozen OpenVLA checkpoint fine-tuned on ten
LIBERO-Spatial tasks~\cite{openvla,libero}, using
$224\times224$ RGB observations and language instructions.
Each rollout starts with ten settling actions, excluded
from action counts, followed by up to 220 policy actions;
execution ends upon success or budget exhaustion.
Using LIBERO-Plus categories and assets~\cite{liberoplus}
under our sampling protocol, we collect 50 ID cases and
20 OOD cases per category per task. The seven categories
are Background, Camera, Language, Light, Noise, Layout,
and Robot. Initializations are selected without consulting
outcomes. Binary detection uses all 1,900 initial inputs;
type classification uses the 1,400 OOD inputs.
Offline risk evaluation uses the corresponding OOD rollouts:
751 successes and 649 failures.

\par\smallskip
\noindent\textbf{Implementation and evaluation.}
We use scikit-learn logistic regression
with $L_2$ regularization, $C=1$, and tolerance $10^{-4}$.
The binary classifier leaves image features unstandardized
and weights standardized text features by 0.005, fixed
during development. It uses LIBLINEAR with 4,000 iterations;
the type and risk classifiers use L-BFGS with 3,000 iterations.
Both initial-input classifiers use class-balanced weights.
Our predictor combines seven type indicators,
14 action-prefix features, and 32 PCA components of the
cumulative mean execution-step representation from step 10
through the current step. Representation standardization
and PCA are fitted to these means at up to eight evenly
spaced eligible prefixes per training rollout.
Risk training uses every prefix from $k=10$ to $T-1$,
with equal total weight per rollout in input standardization
and classifier fitting.

We use ten-fold leave-one-group-out cross-validation,
holding out one task per fold. All preprocessing is fitted
on training tasks; training-side type predictions are
generated by further holding out each training task.
We pool held-out predictions and report ROC-AUC, TPR,
and FPR for OOD detection, accuracy and macro-F1 for type
classification, and ROC-AUC, TPR, FPR, and F1 for failure
prediction. Decision thresholds are $d(\boldsymbol{z}_0)>0.5$
and $\hat p_k\geq0.5$, respectively.
Risk evaluation starts at step 10 on all active OOD
rollouts, independently of the binary gate.

\par\smallskip
\noindent\textbf{Baselines.}
We retrain ActProbe~\cite{actprobe2026}, SAFE~\cite{safe},
and logistic context variants on the same rollouts,
outer task splits, labels, and checkpoints.
Within each outer fold, eight tasks train candidates and
one selects configurations by step-60 ROC-AUC before
refitting on all nine tasks.
The action/context logistic baselines select
$C\in\{0.1,1,10\}$; our predictor uses fixed $C=1$.
ActProbe retains its OpenVLA action features, time index,
and instruction-conditioned recurrent architecture,
using execution-side rather than pre-transform commands.
It shares 1,024-dimensional instruction embeddings with
the instruction-context logistic baseline.
SAFE uses First features with its original MLP/LSTM
losses and readouts, including cumulative MLP scores.
Neural baseline ROC-AUC values are averaged over three seeds.

\par\smallskip
\noindent\textbf{Hardware.}
Timing uses an RTX 4090 D GPU and AMD EPYC 9754 CPU
with an 18-vCPU quota, four PyTorch threads, and one BLAS
thread. OpenVLA uses BF16, SDPA, and Transformers 4.40.1;
neural baselines use FP32. Both environments use
PyTorch 2.2.0+cu118, with scikit-learn 1.5.1 for logistic
models. We time warmed-up batch-one calls with inputs
and models in memory, synchronizing GPU operations
when applicable.

\begin{table}[!t]
    \centering
    \vspace{0pt}
    \setlength{\abovecaptionskip}{0pt}
    \setlength{\belowcaptionskip}{2pt}
    \caption{Results for the \textbf{OOD Characterizer}.}
    \label{tab:ood_characterization}

    \resizebox{\linewidth}{!}{%
        \setlength{\tabcolsep}{3pt}
        \renewcommand{\arraystretch}{1.05}
        \begin{tabular}{|l|cc||cc|}
            \hline\thickhline
            \rowcolor{mygray}
            \multicolumn{3}{|c||}{ID/OOD detection}
            & \multicolumn{2}{c|}{OOD type classification} \\
            \hline
            \rowcolor{mygray}
            \multicolumn{1}{|c|}{ROC-AUC}
            & TPR (\%) & FPR (\%)
            & Accuracy (\%) & Macro-F1 \\
            \hline\hline
            \multicolumn{1}{|c|}{0.9454}
            & 84.57 & 6.20 & 91.00 & 0.9123 \\
            \hline\hline

            \rowcolor{mygray}
            OOD type
            & \multicolumn{2}{c||}{Detection TPR (\%)}
            & \multicolumn{2}{c|}{Type accuracy (\%)} \\
            \hline\hline
            Background
            & \multicolumn{2}{c||}{91.00}
            & \multicolumn{2}{c|}{94.50} \\
            Camera
            & \multicolumn{2}{c||}{100.00}
            & \multicolumn{2}{c|}{97.00} \\
            Language
            & \multicolumn{2}{c||}{85.00}
            & \multicolumn{2}{c|}{100.00} \\
            Light
            & \multicolumn{2}{c||}{72.50}
            & \multicolumn{2}{c|}{83.50} \\
            Noise
            & \multicolumn{2}{c||}{89.00}
            & \multicolumn{2}{c|}{90.00} \\
            Layout
            & \multicolumn{2}{c||}{61.00}
            & \multicolumn{2}{c|}{84.50} \\
            Robot
            & \multicolumn{2}{c||}{93.50}
            & \multicolumn{2}{c|}{87.50} \\
            \hline
        \end{tabular}%
    }

    \par\smallskip
    \begin{minipage}{\linewidth}
        \footnotesize\raggedright
        Type classification is evaluated on all OOD cases, independently of the gate.

    \end{minipage}
\end{table}

\subsection{OOD Characterizer Results}

Table~\ref{tab:ood_characterization} reports an OOD detection
ROC-AUC of 0.9454 and type accuracy of 91.00\%.
Detection TPR varies from 61.00\% for Layout to 100.00\%
for Camera. All 200 Language cases receive correct type
labels, yet only 170 pass the binary gate, showing that
correct type classification does not ensure admission
to the risk predictor.
An ID-only cosine 1-nearest-neighbor baseline achieves
ROC-AUC 0.8176 on the same 1,900 held-out inputs, using
$1-\max$ cosine similarity over 450 training ID
representations per fold. These representations are
L2-normalized in the original 8,192-dimensional space
without text weighting. The supervised classifier
achieves higher ROC-AUC, but also uses OOD training
examples unavailable to the distance baseline.

\subsection{Failure Risk Predictor Results}

Failure rates differ across OOD categories, ranging from
27.5\% for Layout to 77.5\% for Camera, with 200 rollouts
per category (Fig.~\ref{fig:intro_motivation}).
Every category contains both successful and failed
rollouts. Thus, shift type provides context about
outcome frequency under the sampled perturbations,
but does not determine whether an individual execution
will fail.

On all 1,400 OOD cases, ROC-AUC increases from 0.6852
at step 10 to 0.7675 at step 30 and 0.8497 at step 60
(Table~\ref{tab:risk_fixed_steps}). Because the cohort
remains unchanged at these checkpoints, this improvement
supports the value of accumulating execution evidence.
Later results also reflect changes in the evaluated
cohort as completed rollouts leave evaluation; only
seven successes remain at step 200.

By step 60, 542/649 failures and 381/751 successes have
crossed 0.5 at least once, yielding 83.51\% recall and
a 50.73\% false-alarm fraction. In contrast, the current
step-60 decision yields 70.72\% TPR and 12.52\% FPR.
An ever-alarm rule detects more eventual failures,
but also records temporary warnings during successful
executions.

On the 1,184 gate-admitted OOD cases, step-60 ROC-AUC
is 0.8534, compared with 0.8497 on all OOD cases.
This conditional result excludes 216 cases, including
45 eventual failures that receive no risk score under
gated operation.

Table~\ref{tab:risk_progress} retains all 1,400 cases
at retrospective checkpoints derived from final rollout
lengths. Action counts are rounded down, with 100\%
evaluated at $T-1$. ROC-AUC increases from 0.8329 at
25\% progress to 0.9818 at the final scored prefix,
showing stronger outcome separation on a fixed cohort
as execution proceeds.

Table~\ref{tab:risk_baselines} compares our predictor
with ActProbe and SAFE. Our ROC-AUC exceeds all three
evaluated baselines at both checkpoints. The gap over
ActProbe is small at step 30 (0.0074) and larger at
step 60 (0.0365). Thus, the observed advantage is more
pronounced after a longer execution prefix, rather than
uniform across checkpoints. CPU scoring latency is
also lower at both checkpoints under the cached-input
timing protocol.

\begin{table}[t]
    \centering
    \setlength{\abovecaptionskip}{0pt}
    \setlength{\belowcaptionskip}{2pt}
    \caption{\textbf{Failure-risk results} at selected
    executed-action checkpoints.}
    \label{tab:risk_fixed_steps}
    \small

    \resizebox{\columnwidth}{!}{%
        \setlength{\tabcolsep}{2pt}
        \renewcommand{\arraystretch}{1.05}
        \begin{tabular}{|r|rr||rrrr|}
            \hline\thickhline
            \rowcolor{mygray}
Step & S & F & ROC-AUC $\uparrow$ & TPR (\%) $\uparrow$ & FPR (\%) $\downarrow$ & F1 (\%) $\uparrow$ \\
            \hline\hline
10 & 751 & 649 & 0.6852 & 59.94 & 36.09 & 59.43 \\
30 & 751 & 649 & 0.7675 & 58.09 & 13.85 & 66.73 \\
60 & 751 & 649 & 0.8497 & 70.72 & 12.52 & 76.37 \\
90 & 598 & 649 & 0.9108 & 80.12 & 11.87 & 83.87 \\
120 & 226 & 649 & 0.9235 & 86.44 & 15.93 & 90.05 \\
160 & 25 & 649 & 0.8395 & 92.76 & 52.00 & 95.25 \\
200 & 7 & 649 & 0.8759 & 95.99 & 57.14 & 97.65 \\
219 & 0 & 649 & -- & 96.30 & -- & 98.12 \\
            \hline
        \end{tabular}%
    }

    \par\smallskip
    \begin{minipage}{\columnwidth}
        \footnotesize\raggedright
        S/F: eventual successes/failures still active. Dashes denote undefined metrics.

    \end{minipage}
\end{table}
\begin{table}[t]
    \centering
    \setlength{\abovecaptionskip}{0pt}
    \setlength{\belowcaptionskip}{2pt}
    \caption{Retrospective progress-based evaluation.}
    \label{tab:risk_progress}
    \small
    \resizebox{\columnwidth}{!}{%
        \setlength{\tabcolsep}{3pt}
        \renewcommand{\arraystretch}{1.05}
        \begin{tabular}{|l||rrrr|}
            \hline\thickhline
            \rowcolor{mygray}
        Progress & ROC-AUC $\uparrow$
        & TPR (\%) $\uparrow$
        & FPR (\%) $\downarrow$
        & F1 (\%) $\uparrow$ \\
                    \hline\hline
25\% & 0.8329 & 68.88 & 15.05 & 73.95 \\
50\% & 0.9186 & 84.59 & 13.85 & 84.33 \\
75\% & 0.9699 & 93.37 & 10.52 & 90.85 \\
100\% & 0.9818 & 96.30 & 12.25 & 91.51 \\
        \hline
        \end{tabular}%
    }
    \par\smallskip
    \begin{minipage}{\columnwidth}
        \footnotesize\raggedright
        All 1,400 cases remain; 100\% uses prefix $T-1$.
    \end{minipage}
\end{table}

\begin{table}[t]
\centering
\caption{Failure-risk ROC-AUC summaries and CPU scoring
latency at 30 and 60 executed actions.}
\label{tab:risk_baselines}
\small
\setlength{\tabcolsep}{3pt}
\renewcommand{\arraystretch}{1.15}

\resizebox{\columnwidth}{!}{%
\begin{tabular}{p{3.2cm}rrrr}
    \thickhline
    \rowcolor{mygray}
    & \multicolumn{2}{c}{30 executed actions}
    & \multicolumn{2}{c}{60 executed actions} \\
    \cline{2-3}\cline{4-5}
    \rowcolor{mygray}
    Method
    & \shortstack{ROC-AUC\\$\uparrow$}
    & \shortstack{Latency\\(ms) $\downarrow$}
    & \shortstack{ROC-AUC\\$\uparrow$}
    & \shortstack{Latency\\(ms) $\downarrow$} \\
    \hline
    ActProbe
    & 0.7601 & 0.703 & 0.8132 & 1.068 \\
    SAFE-MLP
    & 0.7294 & 1.712 & 0.8055 & 3.249 \\
    SAFE-LSTM
    & 0.6867 & 4.501 & 0.7414 & 6.773 \\
    \rowcolor{mygray}
    \name{}
    & 0.7675 & 0.608 & 0.8497 & 0.893 \\
    \thickhline
\end{tabular}%
}

\end{table}

\begin{figure*}[!t]
\centering
\includegraphics[width=\textwidth]{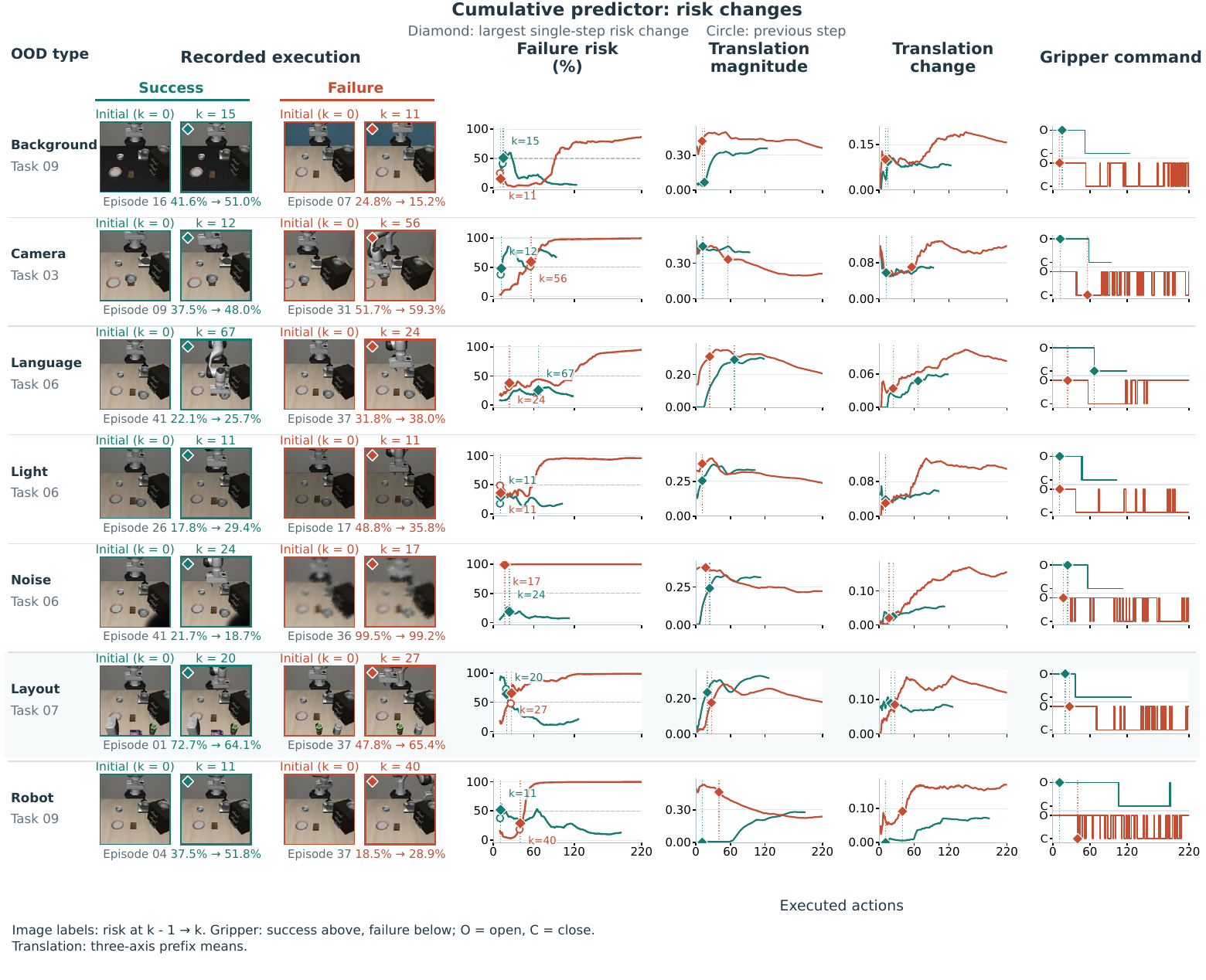}
\caption{Failure Risk Predictor examples across seven OOD categories.
Frames show initialization and the largest single-step risk change;
diamonds mark the selected step and hollow circles the preceding step.}
\label{fig:qualitative_errors}
\end{figure*}

\subsection{Ablation Study}

We retrain reduced-input variants with $C=1$ and the same
task-grouped folds. Table~\ref{tab:module1_input_ablation}
compares initial image and text inputs. Binary text features
retain their standardization and 0.005 scaling; type
classification standardizes the retained features.
Combining image and text improves aggregate detection and
type accuracy over either modality alone. For Language OOD,
detection increases from 18 to 170 of 200 cases compared
with Image only, and correct type labels increase from
157 to 200. Layout detection instead decreases from 133 to 122. These results show that language input adds useful information
for identifying instruction shifts, but does not improve
detection uniformly across categories.

\begin{table}[t]
    \centering
    \caption{Input ablation results for the OOD Characterizer.}
    \label{tab:module1_input_ablation}
    \small
    \setlength{\tabcolsep}{3pt}
    \renewcommand{\arraystretch}{1.15}

    \resizebox{\columnwidth}{!}{%
    \begin{tabular}{@{}lrrrrr@{}}
        \thickhline
        \rowcolor{mygray}
        & \multicolumn{3}{c}{\textbf{ID/OOD detection}}
        & \multicolumn{2}{c}{\textbf{OOD type classification}} \\
        \cline{2-4}\cline{5-6}
        \rowcolor{mygray}
        \textbf{Input}
        & \shortstack{\textbf{ROC-AUC}\\$\uparrow$}
        & \shortstack{\textbf{TPR}\\(\%) $\uparrow$}
        & \shortstack{\textbf{FPR}\\(\%) $\downarrow$}
        & \shortstack{\textbf{Accuracy}\\(\%) $\uparrow$}
        & \shortstack{\textbf{Macro-F1}\\$\uparrow$} \\
        \hline
        Image only
        & 0.8827 & 75.00 & 6.20 & 88.14 & 0.8859 \\
        Text only
        & 0.5714 & 31.43 & 20.00 & 27.14 & 0.2379 \\
        Image + Text
        & 0.9454 & 84.57 & 6.20 & 91.00 & 0.9123 \\
        \thickhline
    \end{tabular}%
    }

    \par\smallskip
    \begin{minipage}{\columnwidth}
        \footnotesize\raggedright
        Fixed preprocessing and $C=1$ for each input variant.
    \end{minipage}
\end{table}
Table~\ref{tab:risk_ablation} separates our predictor's input ablations
from the action/type predictor's feature ablations.
Removing the cumulative representation reduces step-60
ROC-AUC from 0.8497 to 0.7906, while removing action
features reduces it to 0.8344. Both variants retain
predicted type, supporting the complementary value of
action features and cumulative representations.
Compared with the variant without action features,
the full model reduces FPR from 16.64\% to 12.52\%
and improves F1 from 75.36\% to 76.37\%, although TPR
decreases from 72.11\% to 70.72\%.
Within the action/type predictor, removing translation
change causes the largest ROC-AUC decrease among the
three action-feature removals, whereas removing gripper
features slightly improves performance at this checkpoint. Thus, although gripper switching is more frequent in failed
rollouts overall, it does not provide an additional ROC-AUC
benefit in this action/type configuration at step 60.

\begin{table}[t]
    \centering
    \setlength{\abovecaptionskip}{0pt}
    \setlength{\belowcaptionskip}{2pt}
    \caption{Ablation results at 60 executed actions.}
    \label{tab:risk_ablation}
    \small
    \resizebox{\columnwidth}{!}{%
        \setlength{\tabcolsep}{3pt}
        \renewcommand{\arraystretch}{1.15}
        \begin{tabular}{|l||rrrr|}
            \hline\thickhline
            \rowcolor{mygray}
            Model & ROC-AUC & TPR (\%) & FPR (\%) & F1 (\%) \\
            \hline\hline
            \rowcolor{mygray}
            \multicolumn{5}{|l|}{Failure Risk Predictor} \\
            \hline
            w/o Cumulative representation
            & 0.7906 & 63.17 & 14.65 & 70.15 \\
            w/o Action features
            & 0.8344 & 72.11 & 16.64 & 75.36 \\
            \rowcolor{mygray}
            Full model (\name{})
            & 0.8497 & 70.72 & 12.52 & 76.37 \\
            \hline\hline
            \rowcolor{mygray}
            \multicolumn{5}{|l|}{Action/type predictor} \\
            \hline
            Actions only
            & 0.7756 & 63.33 & 13.98 & 70.56 \\
            Type only
            & 0.6739 & 58.55 & 24.77 & 62.55 \\
            w/o Translation magnitude
            & 0.7532 & 60.25 & 17.58 & 66.72 \\
            w/o Translation change
            & 0.7064 & 58.86 & 25.03 & 62.67 \\
            w/o Gripper features
            & 0.7929 & 65.18 & 13.58 & 72.06 \\
            \rowcolor{mygray}
            Actions + Type
            & 0.7906 & 63.17 & 14.65 & 70.15 \\
            \hline
        \end{tabular}%
    }
    \par\smallskip
    \begin{minipage}{\columnwidth}
        \footnotesize\raggedright
        Type denotes the predicted OOD category.
        All variants use $C=1$.
    \end{minipage}
\end{table}

Under the selected-$C$ context comparison, step-60 ROC-AUC
is 0.7750 with actions alone, 0.6567 with instruction context,
and 0.7902 with predicted-type context. Predicted-type context
ranks higher within seven of ten tasks than actions alone
and within eight than instruction context.

\subsection{Computational Cost}

For CPU scoring, we select 70 cases by fixed identifier hashes
within task--category strata. For each fold and checkpoint,
we perform ten warm-up calls, followed by ten timed
repetitions per selected case. Scoring includes action features,
recomputing the mean of representations from steps 10 to $k$,
standardization, PCA, concatenation, and classification.
It takes 0.608\,ms at step 30 and 0.893\,ms at step 60
(Table~\ref{tab:risk_baselines}). Raw representations and
context are cached; model loading, file access, simulation,
and policy feature acquisition are excluded. Neural timings
use seed-0 models and recompute recurrent states.

Across 98 initial inputs with five paired repetitions,
adding initial representation capture, pooling, and gated
Module 1 classification increases policy-generation time
by 2.15\,ms (178.80 to 180.94\,ms). SAFE capture adds
0.56\,ms per decision on 56 inputs with five paired repetitions.
These acquisition measurements are separate from CPU scoring.

\subsection{Qualitative Analysis}

Across adjacent scored steps, cumulative averaging reduces
the mean absolute risk change from 3.52 to
0.70 percentage points compared with the
current-step fusion predictor, without smoothing output scores.

Figure~\ref{fig:qualitative_errors} shows our predictor
on success and failure examples across
seven OOD categories. For each rollout, we select the
largest absolute change between consecutive risk scores
and align the corresponding frame and action features.
In the Layout failure example, risk rises from 47.8\%
to 65.4\% at step 27. In the Background failure example,
the selected change is instead a decrease from 24.8\%
to 15.2\% at step 11, followed by a later rise.
These examples show that the largest local risk change
need not be an increase or a threshold crossing.

For additional error analysis, we define false positives
and false negatives at step 60 using a threshold of 0.5,
selecting one case from each error class by a fixed
identifier hash. In the Light false positive (Task 06, Episode 07), the robot briefly pauses during grasping but secures the object and completes the task at step 103, despite a step-60 risk
of 51.59\%. In the Background false negative (Task 02,
Episode 15), movements are small, with no obvious
execution anomaly at step 60, when risk is 42.13\%.
The robot subsequently shows little visible motion
over approximately steps 68--180 and does not complete
the task within the 220-action budget. This case
illustrates a checkpoint prediction that misses eventual
budget-limited failure despite apparently uneventful
early execution.

Across all 1,400 cases at step 60, failure/success median
translation change is 0.1005/0.0743 and gripper switching
is 3.39\%/1.69\%. All 70 task--category strata contain both
outcomes. Failure medians are higher for translation change
in 58 strata; gripper switching is higher in 31, tied in 29,
and lower in 10.

We observe a mismatch between action selection and actual
task progress: a policy may issue lifting or transport commands
as if a grasp had succeeded, even when the object is not secured,
or repeat corrections without achieving the intended state
change. Action features may capture these consequences;
predictor contributions describe changes in its score rather
than physical failure causes.

\section{Conclusion}
We presented \name{}, a two-stage framework that characterizes initial input shifts and predicts failure from partial executions. It combines the predicted shift category with action-prefix and execution progress features while keeping the OpenVLA policy frozen. Using leave-one-task-out cross-validation on ten LIBERO-Spatial tasks, OOD detection achieves a ROC-AUC of 0.9454, and shift classification achieves 91\% accuracy across seven OOD categories. Evaluated independently of the initial OOD gate on all 1,400 OOD rollouts, the failure predictor achieves a ROC-AUC of 0.8497 after 60 executed actions. Adding execution progress features improves ROC-AUC from 0.7906 to 0.8497, an absolute gain of 0.0591 over the predictor using only shift context and action features. At both 30 and 60 actions, the combined predictor achieves higher ROC-AUC and lower CPU scoring latency than the evaluated ActProbe and SAFE baselines. These results demonstrate the value of combining action behavior with execution progress features for accurate and efficient failure prediction under input shifts.

\bibliographystyle{IEEEtran}
\bibliography{main}

\end{document}